\documentclass[a4paper,fleqn]{cas-dc}

\usepackage{caption}

\usepackage[numbers]{natbib}
\def\tsc#1{\csdef{#1}{\textsc{\lowercase{#1}}\xspace}}
\tsc{WGM}
\tsc{QE}
\begin{document}
\let\WriteBookmarks\relax
\def\floatpagepagefraction{1}
\def\textpagefraction{.001}
%\let\printorcid\relax % 可去掉页面下方的ORCID(s)

\shorttitle{}    

% \shortauthors{<short author list for running head>}
%\shortauthors{V. {{\=A}}nand Rawat et al.}

\title[mode = title]{Decoupling SQL Query Hardness Parsing for Text-to-SQL}

\author[]{Jiawen Yi}
[type=editor,
%    auid=000,bioid=1,
%    prefix=Sir, 
%    role=Researcher, 
%    orcid=0000-0001-7511-2910
    ]
%\cormark[2] 
\ead{214601043@csu.edu.cn}
%\ead{cvr_1@tug.org.in} 
%\ead[url]{www.cvr.cc,www.tug.org.in}
%\credit{Conceptualization of this study, Methodology, Software}

\author[]{Guo Chen}
[type=editor,
orcid=0000-0001-5459-4033]
\ead{guochen@ieee.org}
\cormark[1] 
%\author[2,3]{T. Rishi Nair}[role=Co-ordinator, suffix=Jr]
%\ead{rishi@sayahna.org}
%\ead[URL]{www.sayahna.org}
%\credit{Data curation, Writing - Original draft preparation}
%
%\author[1,3]{Karl Berry}
%\cormark[2] 
%\fnmark[1,3]
%\ead{karl@freefriends.org} 
%\ead[URL]{www.tug.org}

\address[]{Departent of Automation, Central South University, Hunan, Changsha, 410083, China}
%\address[2]{Sayahna Foundation, Jagathy, Trivandrum 695014, India}
%\address[3]{\TeX{} Users Group, Providence, MA, USA}

\cortext[1]{Corresponding author} 
%\cortext[2]{Principal corresponding author} 

\begin{abstract}
The fundamental goal of the Text-to-SQL task is to translate natural language question into SQL query. Current research primarily emphasizes the information coupling between natural language questions and schemas, and significant progress has been made in this area. The natural language questions as the primary task requirements source determines the hardness of correspond SQL queries, the correlation between the two always be ignored. However, when the correlation between questions and queries was decoupled, it may simplify the task. In this paper, we introduce an innovative framework for Text-to-SQL based on decoupling SQL query hardness parsing. This framework decouples the Text-to-SQL task based on query hardness by analyzing questions and schemas, simplifying the multi-hardness task into a single-hardness challenge. This greatly reduces the parsing pressure on the language model. We evaluate our proposed framework and achieve a new state-of-the-art performance of fine-turning methods on Spider dev.  

%the importance of the correlation between natural language questions and SQL queries has been largely overlooked. 
\end{abstract}


%\item highlight-1
%\item highlight-2
%\item highlight-3

% Keywords

\maketitle

% Main text
\section{Introduction}

%Text of section-1 \cite{Fortunato2010}.
In the current era, marked by an unprecedented influx of data, elements such as text, integers, and floating-point numbers are ubiquitous. Central to managing this diverse data are relational databases, which underpin modern data management systems. Nonetheless, a salient limitation lies in their restricted usability for laypersons, primarily due to the intricacies of Structured Query Language (SQL). This challenge has catalyzed the development of the Text-to-SQL task, an innovative methodology designed to seamlessly transform natural language inquiries into SQL queries. Historically, the benchmarks for this research endeavor have witnessed a progressive rise in sophistication, transitioning from domain-specific frameworks like ATIS \cite{1} and GeoQuery \cite{2} to more complex, cross-domain benchmarks exemplified by WikiSQL \cite{3} and Spider \cite{4}. Contemporary focus gravitates towards multi-domain, single-turn Text-to-SQL tasks utilizing the Spider benchmark, distinguished for its comprehensive suite of intricate SQL operators (such as ORDER BY, GROUP BY, and HAVING, among others) and nested SQL queries.

In recent years, as researchers have increasingly turned their focus to the Spider dataset, a plethora of methodologies have demonstrated commendable performance. For instance, Bogin et al. \cite{5} introduced schema encoding to augment the generalization capability of the LSTM encoder-decoder framework for database schema generation. Both RASAT \cite{6} and RATSQL \cite{7} emphasized schema linking, a process that aligns entity name mentions within the question to their corresponding schema tables or columns. Concurrently, works like SADGA \cite{8} and LGESQL \cite{9} have delved into the potential of graph neural networks in extracting semantic features from both questions and schemas.

Pre-training language models (PLMs) have reshaped this landscape significantly, facilitating breakthroughs in the Text-to-SQL domain through the deployment of seq2seq PLMs, which have rapidly become a dominant research approach. The mechanism involves feeding the serialized schema items (table names and column names) alongside the question into a seq2seq PLM, utilizing potent tools such as BERT \cite{10}, BART \cite{11}, or T5 \cite{12} to generate the SQL query, as show in Figure 1. Recent noteworthy work included the innovative PICARD by Scholak et al. \cite{13}, employing incremental parsing to curtail the auto-regressive decoder of the language model, effectively mitigating invalid code generation. A surge in efforts to hone domain generalization has also been observed, with pioneering projects like Graphix-T5 \cite{14} incorporating graph-like elements into the Text-to-SQL parser to augment the inferential prowess of PLMs, thereby elevating the efficiency in fabricating complex SQL queries.

In the afore-mentioned research work based on seq2seq PLM, researchers have focused on studying the semantic alignment relationships between natural language questions and all schema items. However, they overlooked that an excessive number of schema items might detract attention from crucial schema elements, potentially leading to suboptimal SQL generation outcomes. Li et al. \cite{15} have a different perspective. They discovered that by decoupling the analysis of schema linking and skeleton, the workload of schema linking during SQL parsing can be alleviated. They proposed RESDSQL and achieved the State-of-the-Art (SOTA) performance at that time on the Spider datasets, creating a significant performance gap with other algorithms. The advent of RESDSQL has paved a new path for subsequent Text-to-SQL researchers – decoupling, simplifying the complex.

However, existing work tends to concentrate on the relationship between natural language questions and schema items, inadvertently neglecting the study of the relationship between natural language questions and SQL queries. For SQL queries complexity analysis, the Spider benchmark \cite{4} recognizing the intricate nature of SQL sample data, formally categorizes SQL queries into four levels of hardness: \textit{Easy}, \textit{Medium}, \textit{Hard}, and \textit{Extra-Hard}. The more complex the SQL query, the higher the demand for parsing capability of language model and the higher training cost. Ideally, this issue could be addressed by exponentially increasing the model size to enhance its parsing capability. In real life, when the model size is constrained by objective conditions (such as expensive and limited GPU memory), it can enhance the performance of model in generating complex SQL queries by intensifying the training on complex SQL query samples (e.g., sample augmentation). However, this approach might reach the upper limit of the model's comprehension capacity, thereby reducing its performance in generating simple SQL queries. Consequently, resolving the contradiction between the generation performance of SQL queries with different complexities, especially when the computational space (e.g., GPU memory) is constrained by objective factors, has become an important research point in this field. 

\begin{figure}[t]
	\centering
	\includegraphics[width=8cm]{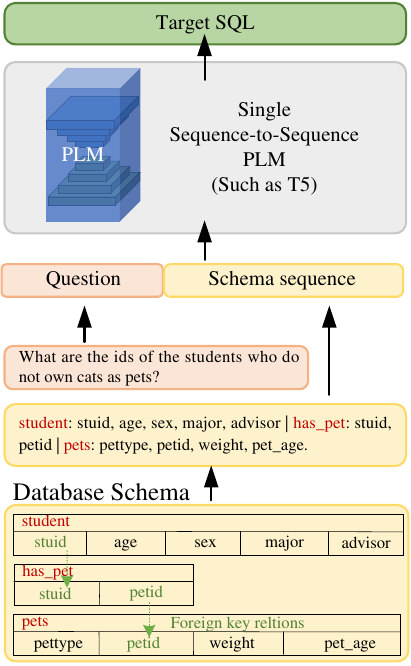}
	\captionsetup[figure]{labelfont={bf}, labelformat={default}, labelsep=period, name={Fig.}}
	\caption{Demonstration of a Text-to-SQL case addressed by a seq2seq PLM. Concatenating natural language question and serialized database schema items (table names and column names) as input, and put it into a  seq2seq PLM to generate target SQL.}\label{fig1}
\end{figure}

In response, we strive to establish a robust linkage between natural language questions and SQL, clarify on a nearly unexplored dimension: the direct correlation between the natural language question semantic and the SQL query hardness (or complexity). Leveraging this insight, we proposed the Decoupling Query Hardness Parsing for Text-to-SQL (DQHP) framework. This innovative architecture disperses parsing tasks across multiple models to simplify the multi-hardness SQL queries generation. It ensures that comprehension capacities of language models are fully leveraged by alleviating the parsing pressure on single language model. This approach maximizes efficiency and cultivates optimal SQL generation performance across various hardness levels. Specifically, our approach utilizes a hardness recognizer to decouple the SQL queries hardness through the natural language questions semantics, wherein a schema ranker is employed to refines schema items, there by enhancing the performance of the hardness recognizer. Subsequently, distributed generator which built by several language models, trained by two-stage training strategy, generate SQL queries in different hardness levels. It is noteworthy that this approach has validated its effectiveness, achieving a notable execution accuracy of 84.7\% on the Spider development dataset with model sizes constrained to 3B or less.

The main contributions are summarized as follows:
\begin{itemize}
	\item[1)] We investigated the correlation between natural language questions and SQL query hardness, introduced the DQHP framework based on the decoupling concept, which mitigates the challenges inherent to the Text-to-SQL task.
	\item[2)] In NLP tasks constrained by model scale (or memory capacity), the DQHP framework offers an approach that distributes the parsing pressure on a language model into multiple language models. Thereby elevating the overall algorithmic performance.
	\item[3)] We conduct evaluations and show that our frame-work achieves the SOTA performance of fine-turning methods with model sizes constrained to 3B or less on Spider dev set.

\end{itemize}

\section{Related works}

Text-to-SQL tasks can be fundamentally seen as translation tasks. Methods based on the Encoder-Decoder structure are one of the most commonly used approaches. They encode the input question and database schema and then decode them into SQL queries. Currently, Encoders are predominantly based on Sequence and Graph encoders.
% while Decoders primarily follow grammar and execution-guided approaches.
Sequence encoder treats text-to-SQL tasks directly as translation tasks, leveraging the advantage of using Pre-trained Language Models (PLMs) for encoding. PLMs already incorporate language patterns within their parameters after pre-training \cite{27}. Specifically, the question and serialized schema items are input into BERT \cite{10}, BART \cite{11}, or T5 \cite{12} for encoding.
Recognizing the multiple associations between question tokens, schema tables, and column names, this approach treats question tokens and schema items as nodes and relationships as edges, constructing heterogeneous graphs (e.g., RAT \cite{21}, LEGSQL \cite{9}, SADGA \cite{8}, S$^2$SQL\cite{25}). These graphs are then encoded using relation-aware transformer networks or relational graph neural networks like RGCN \cite{28} and RGAT \cite{29}. Some methods, like LEGSQL and RAT-SQL, directly employ PLMs to initialize node representations within the graph. However, graph neural networks (GNN) are limited by the over-smoothing issue \cite{30} and depth constraints, Furthermore, GNNs are unable to incorporate language patterns like PLMs, and their architecture is constrained by relationship design, which hinders model robustness and generalization \cite{31}.

Decoders primarily follow grammar and execution-guided approaches. SQL is a language with strong structural grammar, making the use of grammar in decoding effective in reducing the generation of invalid statements. Various methods have been proposed, including top-down decoders generating a series of predefined actions describing the SQL query's syntax tree \cite{32}. Others, like SQLNet \cite{33} and SQLova \cite{34}, design decoders for decoding specific parts of SQL queries based on the SQL syntax structure. SyntaxSQLNet \cite{35} employs decoders based on SQL-specific syntax trees, incorporating a history of SQL generation paths and column-attention encoders that are aware of table structures. A bottom-up decoder called SmBoP \cite{23}, designed by Rubin and Berant, enhances efficiency through bottom-up parallel decoding. PICARD \cite{13} incorporates an incremental parser into the auto-regressive decoder of the PLM. This parser is utilized to eliminate invalid segments of the generated SQL queries throughout the beam search process.
Wang et al. \cite{36} introduced an execution-guided mechanism to utilize SQL semantics. It adjusts partial program execution during decoding, detecting and excluding erroneous programs during the process. Suhr et al. \cite{37} also check the executability of each candidate SQL query to avoid modifying the decoder, which is a strategy also adopted in our method.

In our proposed method, both the schema ranker and hardness recognizer fundamentally involve classification tasks.
Schema items classification serves as an auxiliary task in the context of text-to-SQL, enhancing the model's sensitivity to the associated features between schema items and question tokens. In earlier pre-trained models, this was achieved by training the model to classify schema items as one of the pre-training objectives. For instance, methods like GRAPPA \cite{21} and GAP \cite{39} employed schema items classification as one of their pre-training tasks. Additionally, schema items classification can be implemented as a multi-task learning approach within the text-to-SQL task, as demonstrated in methods like LGESQL. Moreover, some approaches directly employ PLM to predict the relevance of schema items to question tokens, as seen in RESDSQL \cite{15}.
While text classification is less commonly used in the Text-to-SQL domain, it has matured in other fields. For instance, RoBERTa \cite{10} is often utilized as a PLM for text classification tasks in various domains.

Owing to the significant gap between natural language questions and their corresponding SQL queries, certain researchers concentrate on exploring ways to utilize Intermediate Representation (IR) to alleviate the challenges posed by this gap. RAT-SQL and SyntaxSQL \cite{35} use methods to remove the JOIN ON Clause and FROM clause to implement IR. Guo et al. \cite{38} propose using a syntax-based neural model (IRNet) to synthesize SemQL queries to bridge the gap between the intent expressed in natural language and the implementation details in SQL, addressing challenges posed by numerous out-of-vocabulary words in predicting columns. NatSQL \cite{22} achieves improved performance by further simplifying queries. This involves eliminating operators and keywords like FROM, HAVING, GROUP BY, JOIN ON elements that are frequently difficult to match with equivalents in textual descriptions. It also entails eliminating nested subqueries and set operators, and simplifying schema linking by reducing the necessary number of schema items.

\section{Preliminaries}

\textbf{Text-to-SQL}: The process of the Text-to-SQL task is as follows: Formally, given a natural language question $q$, a database $D$, and its schema $S$, the question $q$ is transformed into an SQL query $l$ through an algorithmic model. This query $l$ can be executed on $D$ to obtain the answer to question $q$.

\textbf{Database Schema}: The relational database $D$ determines the scope of the validity of question $q$. For instance, given a database $D$ whose knowledge domain is $K$, if the objective of question $q$ exceeds domain $K$, then the question cannot be answered correctly. The schema $S$ of a database $D$ consists of the following three parts: (1). A series of tables $T=\{t_1,t_2,t_3,…,t_n\}$; (2). A series of columns in each table $C=\{c_{1},c_{2},…,c_{N}\}$; (3). A series of foreign key relationships $R=\{(c_{ik},c_{jh})|c_{ik},c_{jh} \in C\}$, indicating that the information in columns $c_k$ and $c_h$ in different tables $t_i$ and $t_j$ is related.

\textbf{SQL Hardness}: In this paper, the complexity level of SQL query statements is referred to as SQL hardness. We classify SQL hardness into four categories according to the Spider standard \cite{4}: \textit{Easy}, \textit{Medium}, \textit{Hard}, \textit{Extra-hard}. Spider defines the difficulty by taking into account the count of SQL components, selections, and conditions. Consequently, queries featuring an increased number of SQL keywords (such as ORDER BY, GROUP BY, INTERSECT, column selections and aggregators, nested subqueries, etc.) are deemed more challenging. \cite{4}. Figure 2 shows 4 hardness levels examples of SQL queries. Specifically, we set three counts: countA, countB, and countO. Their computation is as follows:

\textit{countA} = (number of WHERE, groupBy, orderBy clauses in query) + (number of tables involved in FROM -1) + (existence of “limit” $+$1).

\textit{countB} = The number of objects in the nested sub-queries in the query.

\textit{countO} = (when the number of times AGG functions (MAX, MIN, COUNT, SUM, AVG) are used in the SELECT, WHERE, groupBy, orderBy clauses of the query > 1, $+$1) + (when the number of columns in SELECT clause >1, $+$1) + (when the number of conditions in WHERE clause >1, $+$1) + (when the number of clauses in groupBy >1, $+$1).

Using the aforementioned three counts, we define SQL Hardness. The specific rules are shown in Table 1.
		
 \begin{table}[t]
	\caption{The definition rule of SQL hardness levels on Spider \cite{4}. The definition of countA, countB and countO is described in SQL Hardness of Section 2. }\label{tbl1}
	\begin{tabular*}{\tblwidth}{@{}LR@{}}
		\toprule
		Hardness level & Condition \\
		\midrule
		Easy: & (countA <= 1, countB = 0, countO = 0) \\
		\midrule
		Medium: & (countA <= 1, countB = 0, 1 <= countO <= 2) \\
		& OR (1 <= countA <= 2, countB = 0, countO < 2) \\
        \midrule
		Hard: & (countA <= 1, countB = 0, countO > 2) \\
		& OR (2 < countA <= 3, countB = 0, countO > 2) \\
		& OR (2 <= countA <= 3, countB = 0, countO <= 2) \\
		& OR (countA <= 1, countB = 1, countO = 0) \\
		\midrule
		Extra-hard: &  meets all other conditions not mentioned above \\
		\bottomrule
	\end{tabular*}
\end{table}

%Easy: countA <= 1, countB = 0, countO = 0.
%Medium: countA <= 1, countB = 0, 1 <= countO <= 2, or 1 <= countA <= 2, countB = 0, countO <2.
%Hard: countA <= 1, countB = 0, countO > 2, or 2 < countA <= 3, countB = 0, countO > 2, or 2 <= countA <= 3, countB = 0, countO <= 2, or countA <= 1, countB = 1, countO = 0.
%Extra-hard: meets all other conditions not mentioned above.

In the Text-to-SQL task workflow, natural language questions serve as the source of all user requirements. Given a database schema, the requirements within the natural language questions determine the hardness of generating the corresponding SQL. So, the semantic of natural language questions often has a direct correlation with the difficulty of generating the associated SQL query.

 \begin{figure}[t]
	\centering
	\includegraphics[width=8cm]{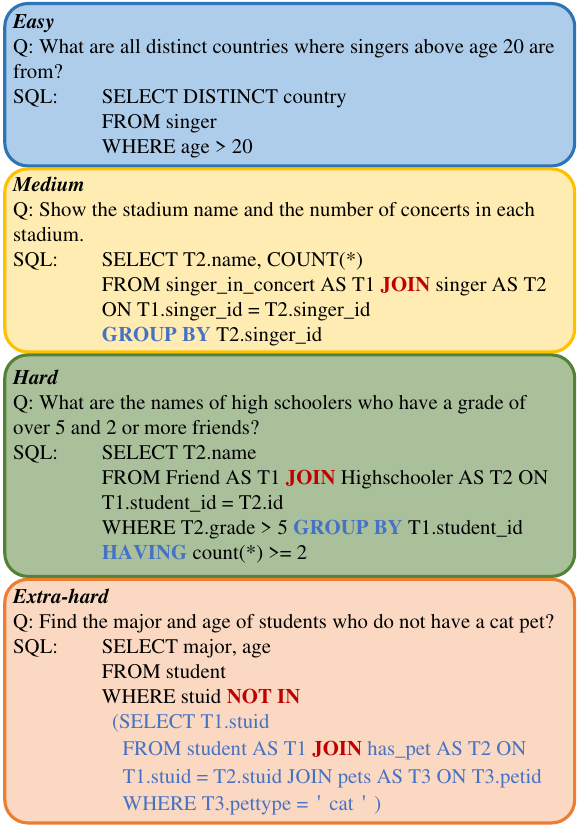}
	\caption{Examples of SQL queries in 4 hardness levels. }\label{fig2}
\end{figure}

%Text of section-2 \cite{Fortunato2010}.

\section{Methodology}

 \begin{figure*}[t]
	\centering
	\includegraphics[width=15cm]{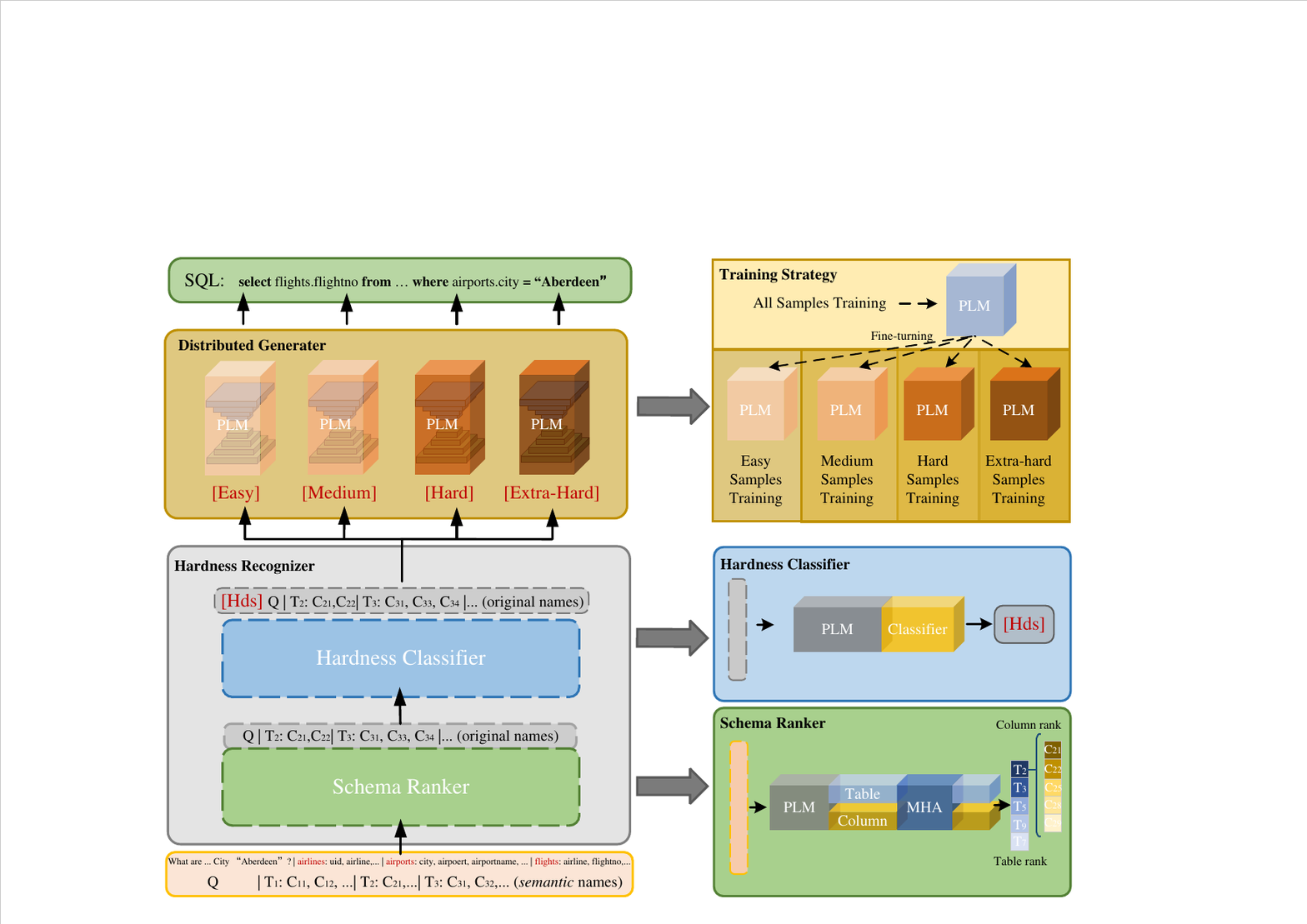}
	\caption{The overview of DQHP framework. We design hardness recognizer to refine the schema items and obtain SQL hardness corresponding question [Hds] for decoupling hardness. Then, distributed generator which training with two-stage training strategy generate SQL query. }\label{fig3}
\end{figure*}

\subsection{Overview}
In the Text-to-SQL task, the hardness of generating SQL queries is largely determined by the natural language question itself. The performance of generating complex SQL queries depends on comprehension capacity of language model, which always is limited by model size. So, how to simplify Text-to-SQL task is a hot issue. 

To address this challenge, we proposed a decoupling SQL query hardness parsing framework, which consists of two main parts: the Hardness Recognizer and the Distributed Generator. This involves first identifying the hardness level of a question through hardness recognition and then utilizing a specialized distributed generator to generate SQL queries. We provide a high-level overview of the proposed DQHP framework in the Figure 3. In the hardness recognizer, we input the question and schema sequence, employing a schema ranker to obtain high-relevance schema items. This procedure effectively reduces the computational burden on subsequent stages of the model. Then, the SQL query hardness determined by hardness classifier corresponding to the question semantics and schema linking. The distributed generator is composed of several independently hardness-specific seq2seq PLMs that trained by two-stage training strategy for enhancing comprehension of each single hardness SQL task. By employing hardness-specialized SQL generation, it aims to distribute the comprehension burden across language models, thereby enhancing the overall SQL generation performance. Through this approach, the task is decoupled from a multi-hardness SQL generation challenge into single-hardness SQL challenge. 

\subsection{Hardness Recognizer}
Ensuring the precision of SQL hardness recognition stands as a fundamental prerequisite and a focal point of SQL hardness decoupling and analysis. However, when the input $X$ includes both the question and the complete set of schema items, directly feeding $X$ into the hardness classifier may lead to obscure or disrupt crucial schema items and the semantic characteristics of the question by large number of schema items. Thereby blurring the perspective of language model and diminishing the effectiveness of SQL hardness classification. Consequently, we introduce a schema ranker to effectively filters and refines the extensive array of schema item information, selecting the most pertinent schema items. These refined schema items are subsequently integrated with the question as inputs to the hardness classifier, thereby optimizing the precision and efficacy of SQL hardness classification.

\textbf{Schema Ranker}. Prior to classifying the SQL hardness corresponding to a given question, we employ a relevance-based schema items selection process, ranking-enhanced encoder of RESDSQL \cite{15} as schema ranker. This process involves extracting schema items that exhibit the highest semantic relevance to the question. The process of schema inputs is designed to alleviate the decoding pressure on the hardness classifier, subsequently enhancing its performance.

In this paper, we employ the common practice that concatenating natural language questions with serialized schema elements as our input format, expressed as $X = [q | t_1: c^1_{1},...,c^{1}_{n_1}|...|t_N: c^{N}_1, ..., c^{N}_{n_N}]$. Here, "|" serves as the delimiter between the question and the schema elements, and between tables, while ":" signifies the relationship between column names and their respective tables. The column names in each table are separated by ",".

Ranking-enhanced encoder \cite{15} as our schema ranker built by a RoBERTa \cite{16} with pooling module and column-enhanced layer. RoBERTa, a refined model derived from BERT \cite{10}, is employed for primary semantic feature extraction from the input sequence $X$, obtain words embedding with rich semantic features. Because of ranking-enhanced encoder using semantic names ("stuid" semantic name is "student id"), PLM's tokenizer might make a mistake on schema items, such as tokenized column name "student id" into "student" and "id" that should be a unit but separated. To avoid this problem, ranking-enhanced encoder using a pooling module that consist of a two-layer Bi-LSTM and a non-linear fully connected layer. Furthermore, for enhancing the model perception of column names, column information is injected into the table embedding by column-enhanced layer based on multi-head scaled dot-product attention layer \cite{17}. Its formulation shows following:
\begin{equation}
	\begin{split}
		T_{c_i}& = MutilHeadAtten(T_i, C_i) \\
		\hat{T_i}& = Norm(T_i+T_{c_i})
	\end{split}
\end{equation}
In the equations provided, $T_i$, $T_{ci}$ and $\hat{T_i}$ represent the table embedding, the column-attentive table embedding and column-enhanced table embedding, respectively. $C_i$ represents the column embedding. $Norm()$ denotes the L2-based regularization function. Then, using two non-linear fully connected layers to achieve top-$k_1$ relevant tables and  top-$k_2$ relevant columns for each table. Therefore, the schema ranker can be formulated as:
\begin{equation}
	\begin{split}
		H_{tc}& =RoBERTa(X) \\
		T,C& =FC(BiLSTM(H_{tc})) \\
		\hat{T}& =CELayer(T,C) \\
		P_t& = FC(\hat{T}) \\
		P_c& = FC(C)
	\end{split}
\end{equation}
where $CELayer()$ and $FC()$ represent the column-enhanced layer function and non-linear fully connection layer function. $H_{tc}$ means hidden state of RoBERTa output. $T$ and $C$ represent tables embedding and columns embedding. $P_t$ and $P_c$ represent prediction of the top-$k_1$ relevant tables and top-$k_2$ relevant columns of each table.

The loss function $L$ designed base on focal loss \cite{18} to solve the imbalanced distribution of positive and negative samples.
\begin{subequations}
	\begin{align}
		 L=\frac{1}{N}\sum_{N}^{i=1}  FL(y_i,\hat{y_i} ) +\frac{1}{M}\sum_{N}^{i=1}\sum_{n_j}^{k=1}  FL(y^i_k,\hat{y} ^i_k ) 	\tag{2}
	\end{align}
\end{subequations}
where $FL$ denotes the focal loss function and $y_i$ is the ground truth label of the $i$-$th$ table. $y^i_k$ is the ground truth label of the $k$-$th$ column in the $i-th$ table. $y_i =1$ or $y^i_k = 1$ indicates the table or column is referenced by the SQL query and 0 otherwise. $\hat{y_i}$ and $\hat{y}^i_k$ are predicted probabilities. 

\textbf{Hardness Classifier}. Upon refining the schema items using the schema ranker, its output is fed into the hardness classifier for hardness recognition. We employ RoBERTa as the language model for encoding semantics of question and schema items, and utilize MLP with a Softmax function as classifiers. By utilizing refined schema items, the encoding process in the RoBERTa language model effectively mitigates the information noise introduced by excessive schema items. This enhancement in encoding quality has notably and significantly improved the classification accuracy for \textit{Medium}, \textit{Hard}, and \textit{Extra-hard} SQL queries. In addition, the training process is performed based on Cross Entropy Loss. 

\subsection{Distributed Generator}
The majority of seq2seq text-to-SQL methods typically employ a single language model for generate any hardness SQL query. However, the comprehension capacity of a single language model is always constrained by its language model size, and increasing the model size demands a larger running memory that potentially resulting in a significant cost increases for computational device. To address the challenge of comprehension capacity, we introduce a Distributed Generator that enhances overall comprehension capacity without increasing the runtime memory requirements. It consists of several independent hardness-specific language models, where each model is exclusively designed for encoding and generating the SQL queries corresponding to a particular hardness level. Additionally, we employ a two-stage training strategy to ensure the comprehension capacity of language model and avoid under-fitting.

\textbf{Distributed Generator}. To address the conflict between the comprehension capacity of language model and runtime memory limitations, we drew inspiration from the concepts of "parallel branching" and "mutual independence" to construct a distributed structure generator. As illustrated in Figure 3, it consists of $h$ independent language models, where $h$ corresponds to the number of categories determined by the hardness recognizer. Each language model in this architecture exclusively receives inputs of the same hardness category, consisting of a question and refined schema items, and then generates the corresponding SQL query. In essence, each language model is specialized in comprehension a specific hardness level of the text-to-SQL task. For instance, in the Spider dataset, this approach decouples the task complexity previously imposed on a single language model, which had to handle all four hardness levels simultaneously, into a language model that only need to address one hardness level. This effectively alleviates the comprehension burden on a single language model. In this approach, there are four language models, each corresponding to a specific hardness category: \textit{Easy}, \textit{Medium}, \textit{Hard}, and \textit{Extra-hard}. Upon determining the hardness level via the hardness recognizer, both the question and the refined schema items are fed into the corresponding language model to the generation of the SQL query. This approach effectively frees up the language model's comprehension capacity from the demands of SQL hardness, allowing the language model to allocate more of its comprehension capacity to single-hardness Text-to-SQL tasks. By fully harnessing these available resources, we can enhance the language model's comprehension abilities and improve the precision of SQL generation. 

\textbf{Two-stage Training Strategy}. In the distributed structure, the approach for dispersing the comprehension burden of a single language model to multiple language models, not only releases the comprehension capacity of each model but also introduces training challenges. Conventional training method may not effectively reinforce a language model comprehension of a single-hardness task. However, due to limited samples quantities, constructing single-hardness training datasets for each language model may lead to under-fitting. Furthermore, Text-to-SQL samples are complex, and data augmentation poses significant challenges and costs.

To address this issue, we adopt a two-stage training strategy, as illustrated in Figure 3. This strategy involves a two-step training process: basic training and specialized training. In the basic training phase, we utilize all hardness samples as the training set to ensure that the language model possesses a fundamental comprehension of the Text-to-SQL task. This basic training results in the creation of a base model, denoted as $Im_i$. Then, the training dataset is divided into four categories based on hardness: \textit{Easy}, \textit{Medium}, \textit{Hard}, and \textit{Extra-hard}. Each language model initialized with the base model $Im_i$, and then specific training with the corresponding hardness training set. The specialized training aims to enhance the model comprehension capabilities for a specific hardness task, enabling each language model to specialize in a particular hardness level. By employing the two-stage training strategy, we not only ensure that the language models possess a foundational comprehension of the Text-to-SQL task but also equip them with specialized expertise in handling specific hardness levels.

%Text of section-3 \cite{Vehlowetal2013}.
\section{Experiments}
\subsection{Experiments Setup}
\textbf{Datasets}. In this paper, we validate our proposed method on the Spider datasets, which is currently one of the most challenging benchmarks for single-turn multi-domain Text-to-SQL tasks. As illustrated in Table 2, the Spider datasets comprises a training set, a development set, and a hidden test set. The training set contains 7,000 samples, with the proportions of \textit{Easy}, \textit{Medium}, \textit{Hard}, and \textit{Extra-hard} samples being 24.2\%, 39.67\%, 20.87\%, and 15.26\%, respectively. The development set consists of 1,034 samples, with proportions of each class being 23.98\%, 43.13\%, 16.83\%, and 16.05\%, respectively. The test set contains 2,147 samples, which have not been publicly disclosed yet. So, all experiments are executed on development set of Spider in the paper.

 \begin{table}[t]
	\caption{Distribution of sample number and its ratio on  training set, development set of Spider datasets. }\label{tbl1}
	\begin{tabular*}{\tblwidth}{@{}LL@{}}
		\toprule
		sample hardness label & sample number \\
		\midrule
		\textbf{Train}:&7000  \\ % Table header row
		Easy&1694 (24.20\%) \\
		Medium&2777 (39.67\%) \\
		Hard&1461 (20.87\%) \\
		Extra-hard&1068 (15.26\%) \\
		\midrule
		\textbf{Dev}:&1034  \\ % Table header row
		Easy&248  (23.98\%) \\
		Medium&446  (43.13\%) \\
		Hard&174  (16.83\%) \\
		Extra-hard&166  (16.05\%) \\
%		\midrule
%		\textbf{Test}:&2147(unpublished)  \\ % Table header row
		\bottomrule
	\end{tabular*}
\end{table}

\textbf{Metric Description}. To assess the performance of the Text-to-SQL parser, we employ two metrics, as proposed by Yu et al. \cite{4} and Zhong, Yu, and Klein \cite{26}: Exact-set-Match accuracy (EM) and EXecution accuracy (EX). EM quantifies whether the predicted SQL query, when converted into a specific data structure, precisely matches the gold SQL query \cite{4}. In contrast, EX compares the execution results of the predicted SQL query with the gold SQL query, making it sensitive to generated values. For the schema ranker, we utilize the Area Under ROC Curve (AUC) to evaluate its performance. 

\textbf{Experimental Setup}. Our experiments were conducted using a cloud-based NVIDIA A100 (80GB) GPU for the 3B model experiments, while all other experiments were carried out on two local devices equipped with NVIDIA GeForce RTX 3090Ti (24GB) GPUs.

\textbf{Parameter Settings}. We divided the training of DQHP-SQL into three stages. In the first stage, we trained the schema ranker. The number of heads $h$ in the column-enhanced layer is set to 8. For optimization, we employed AdamW \cite{19} with a batch size of 8 and a learning rate of 1e-5. In the focal loss, the focusing parameter and the weighted factor were configured as 2 and 0.75, respectively. Additionally, we set $k_1$ and $k_2$ to 4 and 5, respectively. In the second stage, we trained the hardness classifier using Adam with a batch size of 8 and a learning rate of 1e-5 for optimization. For the third stage, we implemented a tow-stage training strategy for the seq2seq model using three T5-based models: Base, Large, and 3B. The two-stage training strategy, we employed Adafactor \cite{20} and fine-tuned them with different batch sizes (bs) and learning rates (lr). Specifically, for DQHP-Base, we used (bs=8, lr=1e-4); for DQHP-Large, we used (bs=4, lr=5e-5); and for DQHP-3B, we used (bs=4, lr=5e-5).

 \begin{table}[t]
	\caption{Exact-set-Match accuracy (EM) and EXecution accuracy (EX) results on Spider dev set(\%). }\label{tbl2}
	\begin{tabular*}{\tblwidth}{@{}LL@{}L}
		%	\begin{tabular*}{\tblwidth}{lcc}
			\toprule
			%		\multirow{2}*{Approach} & \multicolumn{2}{c}{Dev set}  \\
			Approach & EM & EX  \\
			\midrule
			SADGA+GAP\cite{8}&73.1& -  \\ 
			RAT-SQL+GRAPPA\cite{21}&73.4& -  \\
			RAT-SQL+GAP+NatSQL\cite{22}&73.7& 75.0  \\
			SMBOP+GRAPPA\cite{23}&74.7& 75.0  \\
			DT-Fixup-SP+RoBERTa\cite{24}&75.5& -  \\
			LGESQL+ELECTRA\cite{9}&75.1& - \\
			S$^2$SQL+ELECTRA\cite{25}&76.4& -  \\
			%		\midrule
			%		\multicolumn{3}{c}{seq2seq methods}\\
			%		\midrule
			T5-3B \cite{13}&71.5& 74.4  \\
			T5-3B+PICARD \cite{13}&75.5& 79.3 \\
			RASAT+PICARD \cite{6}&75.3& 80.5  \\
			Graphix-T5-3B \cite{14} & 75.6 & 78.2\\
			Graphix-T5-3B+PICARD \cite{14}& 77.1 & 81.0\\
			RESDSQL-3B \cite{15} & 78.0 & 81.8 \\
			RESDSQL-3B+NatSQL \cite{15} & \textbf{80.5} & 84.1  \\
			\midrule
			DQHP-Base& 73.0 & 78.9 \\
			DQHP-Base+NatSQL& 74.8 & 80.9 \\
			DQHP-Large& 75.2 & 80.4  \\
			DQHP-Large+NatSQL& 76.5 & 83.4 \\
			DQHP-3B& 77.0 & 82.1  \\
			DQHP-3B+NatSQL& 79.5 & \textbf{84.7}  \\
			\bottomrule
		\end{tabular*}
	\end{table}

\subsection{Results on Spider}

Table 3 shows the results of various methods on Spider. DQHP-3B achieved a 0.3\% higher EX score on the development set compared to the best baseline without auxiliary. Notably, our DQHP-base outperformed the naked T5-3B by 4.5\%, demonstrating that the decoupling approach can significantly reduce the learning difficulty of the Text-to-SQL task and make full advantage of models comprehension capabilities. On other hands, when combined with NatSQL \cite{22}, an intermediate representation of SQL, DQHP-3B+NatSQL achieved 84.7\% of EX on the development set that resulted in a 2.8\% absolute improvement, and it has outperformed current SOTA of fine-turning method RESDSQL-3B+NatSQL by 0.7\% on EX. This means that DQHP-3B+NatSQL achieve the new SOTA of fine-turning method on Spider development set. The substantial difference between EM and EX is attributed to the strict nature of EM \cite{26}. Given that a single question can have multiple valid SQL queries, the disparity between EM and EX better reflects the model's creativity and comprehension of the essence of the Text-to-SQL task.

\begin{table}[ht]
	\caption{The Execution accuracy on different hardness samples of Spider dev set (\%).}\label{tbl3}
		\begin{tabular*}{\tblwidth}{@{}LC@{}CC@{}CC@{}}
%		\begin{tabular*}{\tblwidth}{lcccc}
			\toprule
			\multirow{2}*{Approach} & \multicolumn{4}{c}{EX} \\
			~& Easy & Medium & Hard & Extra  \\
			\midrule
			T5-Large & 85.5 & 70.9 & 55.2 & 41.6  \\
			T5-3B & 89.5 & 78.3 & 58.6 & 40.1  \\
			T5-3B+PICARD & 95.2 & 85.4 & 67.2 & 50.6 \\
%			RASAT+PICARD &96.0&86.5&67.8&53.6 \\
			Graphix-T5-Large &89.9&78.7&59.8&44.0 \\
			Graphix-T5-3B &91.9&81.6&61.5&50.0 \\
			RESDSQL-Base & 91.9 & 83.6 & 68.4 & 51.8  \\
			RESDSQL-Large & 93.6 & 85.4 & 72.4 & 53.6  \\
			RESDSQL-3B & 94.8 & 87.7 & 73.0 & 56.0  \\
%			RESDSQL-3B+NatSQL & 92.7 & 88.3 &  \textbf{77.6} &  \textbf{66.9}  \\
%			\midrule
%			Our proposed methods \\
			\midrule
			DQHP-Base & 94.0& 84.5& 67.8& 53.0 \\
			DQHP-Large & 94.4& 86.3& 68.4& 56.0  \\
			DQHP-3B & \textbf{95.6}&  \textbf{89.0} & 67.2& \textbf{59.0}  \\
			\midrule
			DQHP-Base+NastSQL & 91.9& 84.3& 71.3& 65.1  \\			
			DQHP-Large+NatSQL & 95.2& 87.7& 73.0& 65.1  \\			
			DQHP-3B+NatSQL &  \textbf{95.9} & 88.1 & \textbf{77.6} & \textbf{66.3} \\
			\bottomrule
		\end{tabular*}
	\end{table}

Table 4 presents the SQL generation performance of various algorithms on each hardness samples in the Spider development set. From the Table 4, it is evident that DQHP-3B achieves the best performance on Easy, Medium and Extra-hard hardness, surpassing the best baseline by 0.8\%, 1.3\% and 3.0\%. Notably, our proposed decoupling approach in DQHP has the most pronounced impact on the performance of the Extra-hard sample category, which has the fewest samples and the highest learning difficulty. Specifically, in terms of SQL generation performance for Extra-hard samples, DQHP-base outperforms T5-3B by a significant margin of 12.9\%, while DQHP-3B surpasses T5-3B by an even more substantial margin of 19.5\%. It performs better than RESDSQL as current the state-of-the-art method which model size less 3B on Spider. Furthermore, NatSQL significantly enhances the generation performance of Hard and Extra-hard samples, specifically yielding improvements of 10.4\% and 7.3\%, respectively. 

However, a conspicuous limitation of DQHP has emerged, specifically its suboptimal generation performance for Hard hardness samples. Although utilizing intermediate representations can substantially mitigate this issue, it remains a significant concern within the DQHP framework. Subsequent experimental analysis on the hardness recognizer revealed that the suboptimal performance of the Hard hardness samples is attributed to the imprecision in hardness classification.

\begin{table}[ht]
	\caption{The classification accuracy of Hardness Recognizer (HR) and HR without schema ranker (HR w/o SR) (\%).}\label{tbl4}
	%	\begin{tabular*}{\tblwidth}{@{}LL@{}LL@{}LL@{}LL@{}}
		\begin{tabular*}{\tblwidth}{lccccc}
			\toprule
			\multirow{2}*{Approach} & \multicolumn{5}{c}{Classification Accuracy} \\
			~& Easy & Medium & Hard & Extra & All  \\
			\midrule
			HR & 88.80 &  \textbf{88.86} &  \textbf{64.83} &  \textbf{87.40} &  \textbf{84.04}\\
			HR w/o SR &  94.89 & 69.42 & 57.39 & 67.56 & 70.50 \\
			\bottomrule
		\end{tabular*}
	\end{table}
	
	\begin{table}[ht]
		\caption{The confusion matrix of the hardness recognizer classification results.}\label{tbl5}
		\begin{tabular*}{\tblwidth}{@{}L|L@{}LL@{}LL@{}LL@{}}
			%		\begin{tabular*}{\tblwidth}{c|cccc}
				\toprule
				\multirow{2}*{Prediction} & \multicolumn{4}{c}{Gold} \\
				~& Easy & Medium & Hard & Extra   \\
				\midrule
				Easy & 222 & 13 & 4 & 0 \\
				Medium & 22 & 414 & 39 & 18  \\
				Hard & 4 & 15 & 114 & 34  \\
				Extra & 0 & 4 & 17 & 114 \\
				\bottomrule
			\end{tabular*}
		\end{table}

\subsection{Hardness Recognizer Results and Analysis}
The hardness recognizer is a crucial component that decouples the analysis based on the query hardness, and its performance directly impacts the subsequent performance of language models in Text-to-SQL. Table 5 records the classification accuracy of the hardness recognizer, from which it can be observed that the classification accuracy for Easy, Medium, and Extra categories are more than 85\%. However, the classification performance for Hard samples may appear less reassuring, that is the reason for DQHP suboptimal performance on Hard samples. In order to further analyze the reason, we record the confusion matrix of the hardness recognizer classification results in the Table 6. From this matrix, it can be observed that out of the Hard samples, 34 were misclassified as Extra-hard, accounting for 20.36\% of the total Hard sample. This is attributed to the ambiguous boundary between adjacent hardness levels. For instance, the SQL complexity of a subset of the Extra-hard samples is relatively low and closely aligns with the complexity of the Hard samples. So, how to address this problem will be our research point in the future.

\textbf{Ablation Study of Schema Ranker}. In Table 5, we present the experimental results for the hardness recognizer without the schema ranker. From these results, it is evident that the inclusion of the schema ranker leads to a significant improvement in the recognition accuracy for the Medium, Hard, and Extra-hard categories. Specifically, there is an enhancement of 21.44\% for Medium, 7.44\% for Hard, and 19.84\% for Extra-hard. Although there's a sacrifice in accuracy for the Easy category, the overall precision has increased by 13.54\%. These outcomes underscore the pivotal value of the schema ranker in the precision of problem hardness decoupling. It also highlights the importance of the decoupling approach in mitigating learning challenges.

\textbf{The Search for Potential}. The impact of decoupling errors resulting from classification mistakes on SQL generation performance cannot be overlooked. In order to elucidate the effects of decoupling errors and the potential of the decoupling approach, we conducted DQHP experiments under the assumption of 100\% accuracy in hardness classification, as documented in Table 7. Our findings suggest that only DQHP-base is marginally impacted by the precision of decoupling, while other DQHP configurations exhibit notable performance enhancements. Specifically, under conditions of 100\% accurate decoupling, DQHP-base+NatSQL witnessed a performance boost of 0.3\%, DQHP-large increased by 1.1\%, DQHP-large+NatSQL by 0.4\%, DQHP-3B by 1.2\%, and DQHP-3B+NatSQL by 0.8\%. From this, it is evident that as the model scale augments, the enhancement attributed to the decoupling methodology also escalates, indicating that decoupling methods harbor greater potential in the context of large-scale language models. In addition, the disparity in Hard sample generation performance between the hypothetical and actual scenarios in DQHP-3B validates that the suboptimal performance of DQHP on "Hard" samples stems from the limited accuracy of classification. This sheds light on a potential avenue for enhancing and refining DQHP in the future.

\begin{table*}[ht]
	\caption{ the EX results of DQHP on the Spider development dataset under both practical classification (with 84.04\% classic accuracy) and ideal classification (with 100\% classic accuracy). The values in parentheses denote the difference in comparison to the practical classification. (\%). }\label{tbl6}
	%	\begin{tabular*}{\tblwidth}{@{}LL@{}LL@{}LL@{}LL@{}}
		\begin{tabular*}{\tblwidth}{lccccc|ccccc}
			\toprule
			\multirow{2}*{Approach} &  \multicolumn{5}{c}{EX with 84.04\% Classic Acc} &\multicolumn{5}{c}{EX with 100\% Classic Acc}  \\
			~& Easy & Medium & Hard & Extra & All  & Easy & Medium & Hard & Extra & All   \\
			\midrule
			DQHP-Base  &94.0& 84.5& 67.8& 53.0 & 78.9  & 94.0 & 84.8(0.3)& 67.2(-0.6)& 53.0 & 78.9  \\
			DQHP-Base+NastSQL & 91.9& 84.3& 71.3& 65.1& 80.9 & 92.3(0.4)& 84.5(0.2)& 71.8(0.5) & 65.6(0.5) &81.2(0.3)\\
			DQHP-Large & 94.4& 86.3& 68.4& 56.0& 80.4  & 96.4(0.2)& 86.6(0.3)& 71.3(2.9)& 56.6(0.6) & 81.5(1.1)   \\
			DQHP-Large+NatSQL& 95.2& 87.7& 73.0& 65.1&83.4  & 94.8(-0.4)& 89.7(2.0)& 74.1(1.1)& 65.7(0.6) & 83.8(0.4)   \\
			DQHP-3B &95.6& \textbf{89.0} & 67.2& 59.0 & 82.0  & 95.6&  \textbf{89.7}(0.7) & 71.3(4.1) & 59.6(0.6) & 83.2(1.2)  \\
			DQHP-3B+NatSQL &\textbf{95.9} & 88.1 & \textbf{77.6} & 66.3 & \textbf{84.7}  &  \textbf{96.4}(0.5) & 88.1 & \textbf{79.3}(1.7) & \textbf{68.7}(2.4) & \textbf{85.5}(0.8)  \\
			\bottomrule
		\end{tabular*}
	\end{table*}

\section{Conclusion}
In this paper, we introduced the DQHP frmework for Text-to-SQL, a method that simplifies learning difficulty of the task by decoupling the SQL query hardness. We began with recognize the SQL query hardness by parsing natural language question, where we employed a schema ranker to refine schema items, enhancing information efficiency while reducing the difficulty of hardness recognition. Next, we used a distributed structure to construct an SQL generator and employed two-stage training strategies to make the models more sensitive to hardness levels. This framework alleviated the parsing pressure of language models and reducing the difficulty of Text-to-SQL by decouples SQL query hardness. Extensive experiments on the Spider dataset have demonstrated the effectiveness of DQHP. Simultaneously, we identified the limitations of DQHP. In the future, we will focus our research on how to recognize hardness with greater accuracy.

\section{Acknowledgments}
We thank Rui Zhang and Tao Yu for answering our some questions about Spider. We also thank Haoyang Li for helpful suggestions.This work is supported by National Naturel Science Foundation of China under Grant No. 62073344.

% Use the style of numbering in square brackets.
% If nothing is used, default style will be taken.
%\begin{enumerate}[a)]
%\item 
%\item 
%\item 
%\end{enumerate}  

%\begin{itemize}
%\item 
%\item 
%\item 
%\end{itemize}  

%\begin{description}
%\item[]
%\item[] 
%\item[] 
%\end{description}  

% Figure
% \begin{figure}[t]
% 	\centering
% 		\includegraphics[width=7cm]{Fig1.pdf}
% 	  \caption{Test}\label{fig1}
% \end{figure}

% \begin{table}[<options>]
% \begin{tabular*}{\tblwidth}{@{}LL@{}}
% \toprule
% \midrule
%  & \\
%  & \\
%  & \\
%  & \\
% \bottomrule
% \end{tabular*}
% \end{table}

% Uncomment and use as the case may be
%\begin{theorem} 
%\end{theorem}

% Uncomment and use as the case may be
%\begin{lemma} 
%\end{lemma}



%\bibliographystyle{cas-model2-names}
\bibliographystyle{elsarticle-num}

%\bibliography{cas-refs.bib}
\bibliography{references.bib}

% % Here goes the biography details.

% % Here goes the biography details.

\end{document}